\documentclass[11pt]{article}

\usepackage[totalwidth=480pt, totalheight=680pt]{geometry}

\usepackage{vmargin}
\setmarginsrb{3 cm}{2.5 cm}{3 cm}{1.5 cm}{0 cm}{1 cm}{0 cm}{1 cm}

\usepackage{graphicx}

\usepackage{url}
\usepackage{bm}
\usepackage{enumitem}
\usepackage{xcolor}
\usepackage{booktabs}
\usepackage{hyperref}

\usepackage{xspace}
\usepackage{color}
\usepackage{algorithm,algorithmicx}
\usepackage[noend]{algpseudocode}
\usepackage{hyperref}
\usepackage{caption}
\usepackage{listings}
\usepackage{courier}
\usepackage{amsthm}
\usepackage{ulem}

\usepackage{amsmath}
\usepackage{parskip}
\usepackage{fancyhdr}
\usepackage{setspace,lipsum}
\usepackage[T1]{fontenc}

\title{A Survey on Deep Reinforcement Learning for Data Processing and Analytics}

\makeatletter
\let\thetitle\@title
\let\theauthor\@author
\let\thedate\@date
\makeatother

\newcommand\blfootnote[1]{%
  \begingroup
  \renewcommand\thefootnote{}\footnote{#1}%
  \addtocounter{footnote}{-1}%
  \endgroup
}

\newcommand{\squishlist}
{
	\begin{list}{$\bullet$}
		{
			\setlength{\itemsep}{0pt}
			\setlength{\parsep}{3pt}
			\setlength{\topsep}{3pt}
			\setlength{\partopsep}{0pt}
			\setlength{\leftmargin}{1.5em}
			\setlength{\labelwidth}{1em}
			\setlength{\labelsep}{0.5em}
		}
	}
	
	\newcommand{\squishend}
	{
	\end{list}
}

\begin{document}

{\centering
{\Large \bfseries \thetitle}\\

\vspace{5mm}
{\large Qingpeng Cai$^{*\dag}$, Can Cui$^{*\dag}$, Yiyuan Xiong$^{*\dag}$\blfootnote{$^*$These authors have contributed equally to this work, and M. Zhang is the contact author.}, Wei Wang$^\dag$, \\ Zhongle Xie$^\S$, Meihui Zhang$^\ddag$} \\

\vspace{5mm}
$^\dag$National University of Singapore \hspace{2mm}
$^\S$ Zhejiang University \hspace{2mm} \\
$^\ddag$Beijing Institute of Techonology \hspace{2mm}

{\scriptsize\{qingpeng, cuican, yiyuan, wangwei\}@comp.nus.edu.sg} \hspace{1mm}
{xiezl@zju.edu.cn} \hspace{1mm} {meihui\_zhang@bit.edu.cn}
}

\pagestyle{fancy}
\vspace{5mm}
\begin{abstract}
Data processing and analytics are fundamental and pervasive. Algorithms play a vital role in data processing and analytics where many algorithm designs have incorporated heuristics and general rules from human knowledge and experience to improve their effectiveness. Recently, reinforcement learning, deep reinforcement learning (DRL) in particular, is increasingly explored and exploited in many areas because it can learn better strategies in complicated environments it is interacting with than statically designed algorithms. Motivated by this trend, we provide a comprehensive review of recent works focusing on utilizing DRL to improve data processing and analytics. First, we present an introduction to key concepts, theories, and methods in DRL. Next, we discuss DRL deployment on database systems, facilitating data processing and analytics in various aspects, including data organization, scheduling, tuning, and indexing. Then, we survey the application of DRL in data processing and analytics, ranging from data preparation, natural language processing to healthcare, fintech, etc. Finally, we discuss important open challenges and future research directions of using DRL in data processing and analytics.
\end{abstract}


\section{Introduction}\label{sec:introduction}

In the age of big data, data processing and analytics are fundamental, ubiquitous, and crucial to many organizations which undertake a digitalization journey to improve and transform their businesses and operations. Data analytics typically entails other key operations such as data acquisition, data cleansing, data integration, modeling, etc., before insights could be extracted. Big data can unleash significant value creation across many sectors such as healthcare and retail~\cite{manyika2011big}. However, the complexity of data (e.g., high volume, high velocity, and high variety) presents many challenges in data analytics and hence renders the difficulty in drawing meaningful insights. To tackle the challenge and facilitate the data processing and analytics efficiently and effectively, a large number of algorithms and techniques have been designed and numerous learning systems have also been developed by researchers and practitioners such as Spark MLlib~\cite{meng2016mllib}, and Rafiki~\cite{wang2018rafiki}.

To support fast data processing and accurate data analytics, a huge number of algorithms rely on rules that are developed based on human knowledge and experience. For example, shortest-job-first is a scheduling algorithm that chooses the job with the smallest execution time for the next execution. However, without fully exploiting characteristics of the workload, it can achieve inferior performance compared to a learning-based scheduling algorithm~\cite{mao2019learning}. Another example is packet classification in computer networking which matches a packet to a rule from a set of rules. One solution is to construct the decision tree using hand-tuned heuristics for classification. Specifically, the heuristics are designed for a particular set of rules and thus may not work well for other workloads with different characteristics~\cite{liang2019neural}. We observe three limitations of existing algorithms~\cite{vamanan2010efficuts, li2018cutsplit}. First, the algorithms are suboptimal. Useful information such as data distribution could be overlooked or not fully exploited by the rules. Second, the algorithm lacks adaptivity. Algorithms designed for a specific workload cannot perform well in another different workload. Third, the algorithm design is a time-consuming process. Developers have to spend much time trying a lot of rules to find one that empirically works.

\begin{figure}[H]

\centering

\includegraphics[width=0.8\textwidth]{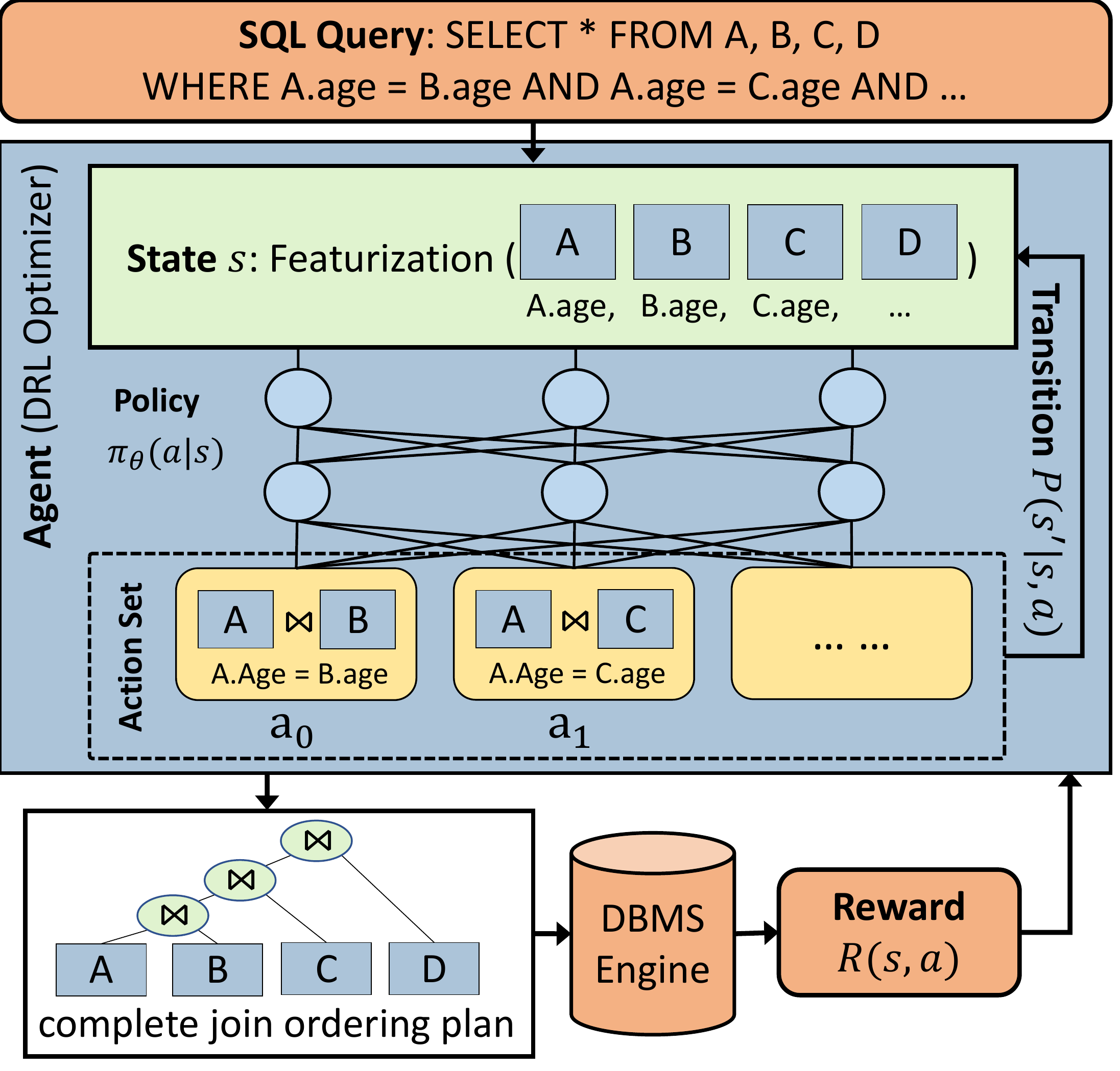}

\caption{The Workflow of DRL for Query Optimization. A, B, C and D are four tables.}

\label{pic:QO}

\end{figure}

Learning-based algorithms have also been studied for data processing and analytics. Two types of learning methods are often used: supervised learning and reinforcement learning. They achieve better performance by direct optimization of the performance objective. Supervised learning typically requires a rich set of high-quality labeled training data, which could be hard and challenging to acquire. For example, configuration tuning is important to optimize the overall performance of a database management system (DBMS)\cite{li2019qtune}. There could be hundreds of tuning knobs that are correlated in discrete and continuous space. Furthermore, diverse database instances, query workloads, hardware characteristics render data collection infeasible, especially in the cloud environment. Compared to supervised learning, reinforcement learning shows good performance because it adopts a trial-and-error search and requires fewer training samples to find good configuration for cloud databases~\cite{zhang2019end}. Another specific example would be query optimization in query processing. Database system optimizers are tasked to find the best execution plan for a query to reduce its query cost. Traditional optimizers typically enumerate many candidate plans and use a cost model to find the plan with minimal cost. The optimization process could be slow and inaccurate~\cite{leis2015good}. Without relying on an inaccurate cost model, deep reinforcement learning (DRL) methods improve the execution plan (e.g., changing the table join orders) by interacting with the database\cite{marcus2018deep, krishnan2018learning}. Figure~\ref{pic:QO} provides a typical workflow for query optimization using DRL. When the query is sent to the agent (i.e., DRL optimizer), it produces a state vector via conducting featurization on essential information, such as the accessed relations and tables. Taking the state as the input, the agent employs neural networks to produce the probability distribution of an action set, where the action set could contain all possible join operations as potential actions. Each action denotes a partial join plan on a pair of tables, and the state will be updated once an action is taken. After taking possible actions, a complete plan is generated, which is then executed by a DBMS to get the reward. In this query optimization problem, the reward can be calculated by the real latency. During the training process with the reward signal, the agent can improve the policy and produce a better join ordering with a higher reward (i.e., less latency).

Reinforcement learning (RL)~\cite{sutton2018reinforcement} focuses on learning to make intelligent actions in an environment. The RL algorithm works on the basis of exploration and exploitation to improve itself with feedback from the environment. In the past decades, RL has achieved tremendous improvements in both theoretical and technical aspects~\cite{silver2018general, sutton2018reinforcement}. Notably, DRL incorporates deep learning (DL) techniques to handle complex unstructured data and has been designed to learn from historical data and self-exploration to solve notoriously hard and large-scale problems (e.g., AlphaGo\cite{Silver_2016}). In recent years, researchers from different communities have proposed DRL solutions to address issues in data processing and analytics\cite{yang2020qd, mao2019learning, 10.1145/3292500.3330868}. We categorize existing works using DRL from two perspectives: system and application. From the system's perspective, we focus on fundamental research topics ranging from general ones, such as scheduling, to system-specific ones, such as query optimization in databases. We shall also emphasize how it is formulated in the Markov Decision Process and discuss how the problem can be solved by DRL more effectively compared to traditional methods. Many techniques such as sampling and simulation are adopted to improve DRL training efficiency because workload execution and data collection in the real system could be time-consuming~\cite{hilprecht2020learning}. From the application's perspective, we shall cover various key applications in both data processing and data analytics to provide a comprehensive understanding of the DRL's usability and adaptivity. Many domains are transformed by the adoption of DRL, which helps to learn domain-specific knowledge about the applications.

In this survey, we aim at providing a broad and systematic review of recent advancements in employing DRL in solving data systems, data processing and analytics issues. In Section 2, we introduce the key concepts, theories, and techniques in RL to lay the foundations. To gain a deeper understanding of DRL, readers could refer to the recently published book~\cite{dong2020deep}, which covers selected DRL research topics and applications with detailed illustrations. In Section 3, we review the latest important research works on using DRL for system optimization to support data processing and analytics. We cover fundamental topics such as data organization, scheduling, system tuning, index, query optimization, and cache management. In Section 4, we discuss using DRL for applications in data processing and analytics ranging from data preparation, natural language interaction to various real-world applications such as healthcare, fintech, E-commerce, etc. In Section 5, we highlight various open challenges and potential research problems. We conclude in Section 6. This survey focuses on recent advancements in exploring RL for data processing and analytics that spurs great interest, especially in the database and data mining community. There are survey papers discussing DRL for other domains. We refer readers to the survey of DRL for healthcare in \cite{yu2019reinforcement}, communications and networking in \cite{luong2019applications}, and RL explainability in \cite{puiutta2020explainable}. Another work\cite{wang2016database} discusses how deep learning can be used to optimize database system design, and vice versa. In this paper, we use "DRL" and "RL" interchangeably.

\section{Theoretical Foundation and Algorithms of Reinforcement Learning}

RL is targeted to solve the sequential decision making problem and the goal is to take actions with maximum expected rewards. In detail, the agent follows a policy to make a series of decisions (i.e. taking actions) in different states of the environment, and the sequence of the states and the actions form a trajectory. To estimate whether the policy is good or not, each decision under the policy will be evaluated by the accumulated rewards through the trajectory. After evaluating the policy from the trajectories, the agent next improves the policy by increasing the probabilities of making decisions with greater expected rewards. By repeating these steps, the agent can improve the policy through trial-and-error until the policy reaches the optimal, and such a sequential decision-making process is modeled via Markov Decision Process (MDP).

\subsection{Markov Decision Process}

\label{subsec:MDP}

Mathematically, MDP, shown in Figure \ref{pic:QO}, is a stochastic control process $\mathcal{M}$ defined by a tuple with 5 elements, $\mathcal{M}=(\mathcal{S},\mathcal{A},\mathcal{R},\mathcal{P},\gamma)$, which are explained as follows.
\begin{itemize}

\item State $\mathcal{S}$: $\mathcal{S}$ is the space for states that denote different situations in the environment and $s_t \in \mathcal{S}$ denotes the state of the situation at the time $t$.

\item Action $\mathcal{A}$: $\mathcal{A}$ is the space for actions that the agent can take; the actions can either be discrete or continuous, and $a_t \in \mathcal{A}$ denotes the action taken at the time $t$.

\item Reward function $\mathcal{R}(s_t,a_t)$: It denotes the immediate reward of the action $a_t$ taken under the state $s_t$.

\item Transition function $\mathcal{P}(s_{t+1} = s'|s_t=s,a_t=a)$: It denotes the probability of transition to the state $s'$ at the time $t + 1$ given the current state $s$ and the taken action $a$ at the time $t$.

\item Discount factor $\gamma \in [0,1]$:
The total rewards of a certain action consist of both immediate rewards and future rewards, and the $\gamma$ quantifies how much importance we give for future rewards.
\end{itemize}

We take the query optimization problem demonstrated in Figure~\ref{pic:QO} to help explain the five components of the MDP. In this example, the state is expressed as a state vector, which summarizes the information of relations and tables that are assessed by the query $q$. In each state, the RL agent produces a probability distribution over all potential actions where each action denotes a partial join plan on a pair of tables. After repeating these two processes, it reaches a terminal state where the final join ordering is generated for an agent to execute, and all actions' target rewards are measured by the actual performance (i.e., latency) or a cost model. As for the transition function, the transitions of the states are always deterministic in both this problem and most of the other DB problems.

In RL, we aim to train the agent with a good policy $\pi$ that is a mapping function from state to action. Through the policy, the agent can take a series of actions that will result in continuous changes in the states, and the sequence of the states and the actions following the policy $\pi$ form a trajectory $\tau = (s_0, a_0, s_1, a_1, ...)$. From each $\tau$, we can evaluate the effect of each action by the accumulated rewards $\mathcal{G}$, and it consists of the immediate reward of this action and the discounted rewards of its following actions in the trajectory. The total result $\mathcal{G}$ for the action $a_t$ is as follows: $\mathcal{G}(\tau) = \sum_{t=0} \gamma^t r_t$, where $\gamma$ quantifies how much importance we give for future rewards. With a bigger $\gamma$, the RL agent will be more likely to take any action that may have a less immediate reward at the current time but has a greater future reward in expectation.

RL continuously evaluates the policy $\pi$ and improves it until it reaches the optimal policy $\pi^* = \arg \max_{(\tau \sim \pi)} \mathcal{G}(\tau)$ where the agent always takes actions that maximize the expected return. To evaluate the policy $\pi$, RL algorithms estimate how good or bad it is for a state and a state-action pair by the function $\mathcal{V}$ and function $\mathcal{Q}$ respectively. Both of these two value functions are calculated according to the discounted return $\mathcal{G}$ in expectation which can be written as:

\begin{eqnarray}
    \mathcal{V}^\pi(s) = E_{\tau \sim \pi} [\mathcal{G}(\tau)|s_0=s] \label{eq:V_expectation} 
\end{eqnarray}
\begin{eqnarray}
    \mathcal{Q}^\pi(s,a) = E_{\tau \sim \pi} [\mathcal{G}(\tau)|s_0=s,a_0=a]
\end{eqnarray}

These two value functions have a close association where the $\mathcal{V}^\pi(s_t)$ is the expectation of the function $\mathcal{Q}$ of all possible actions under the state $s_t$ according to the policy $\pi$, and the $\mathcal{Q}^\pi(s_t,a_t)$ is the combination of the immediate reward of the action $a_t$ and the expectation of all possible states' values after taking the action $a_t$ under the state $s_t$. Hence, we have:
\begin{eqnarray}
    \mathcal{V}^{\pi}(s) &=& \sum_{a \in \mathcal{A}}\pi(a|s)\mathcal{Q}^{\pi}(s,a) \label{eq:V_from_Q} 
    \\
    \mathcal{Q}^{\pi}(s,a) &=& R(s,a) + \gamma \sum_{s' \in \mathcal{S}}\mathcal{P}(s'|s,a)\mathcal{V}^{\pi}(s') \label{eq:Q_from_V}
\end{eqnarray}
Given a policy $\pi$, we can evaluate its value functions by Bellman equations~\cite{sutton2018reinforcement} which utilize the recursive relationships of these value functions. Formally, Bellman equations deduce the relationships between a given state (i.e. function $\mathcal{V}$) or a given state-action pair (i.e. function $\mathcal{Q}$) and its successors which can be written as:

\begin{eqnarray}
    \mathcal{V}^\pi(s) = \sum_{a_t\in \mathcal{A}}\pi(a|s) [R(s,a) + \gamma \sum_{s' \in \mathcal{S}}\mathcal{P}(s'|s,a) \mathcal{V}^\pi(s')]\label{eq:bellman_v}
\end{eqnarray}

\begin{eqnarray}
    \mathcal{Q}^{\pi}(s,a) = \sum_{s' \in \mathcal{S}} \mathcal{P}(s'|s,a) [R(s,a) + \gamma \sum_{a'\in \mathcal{A}}\pi(a'|s')\mathcal{Q}^\pi(s',a')] \label{eq:bellman_q}
\end{eqnarray}

\begin{figure*}

\centering

\includegraphics[width=0.9\textwidth]{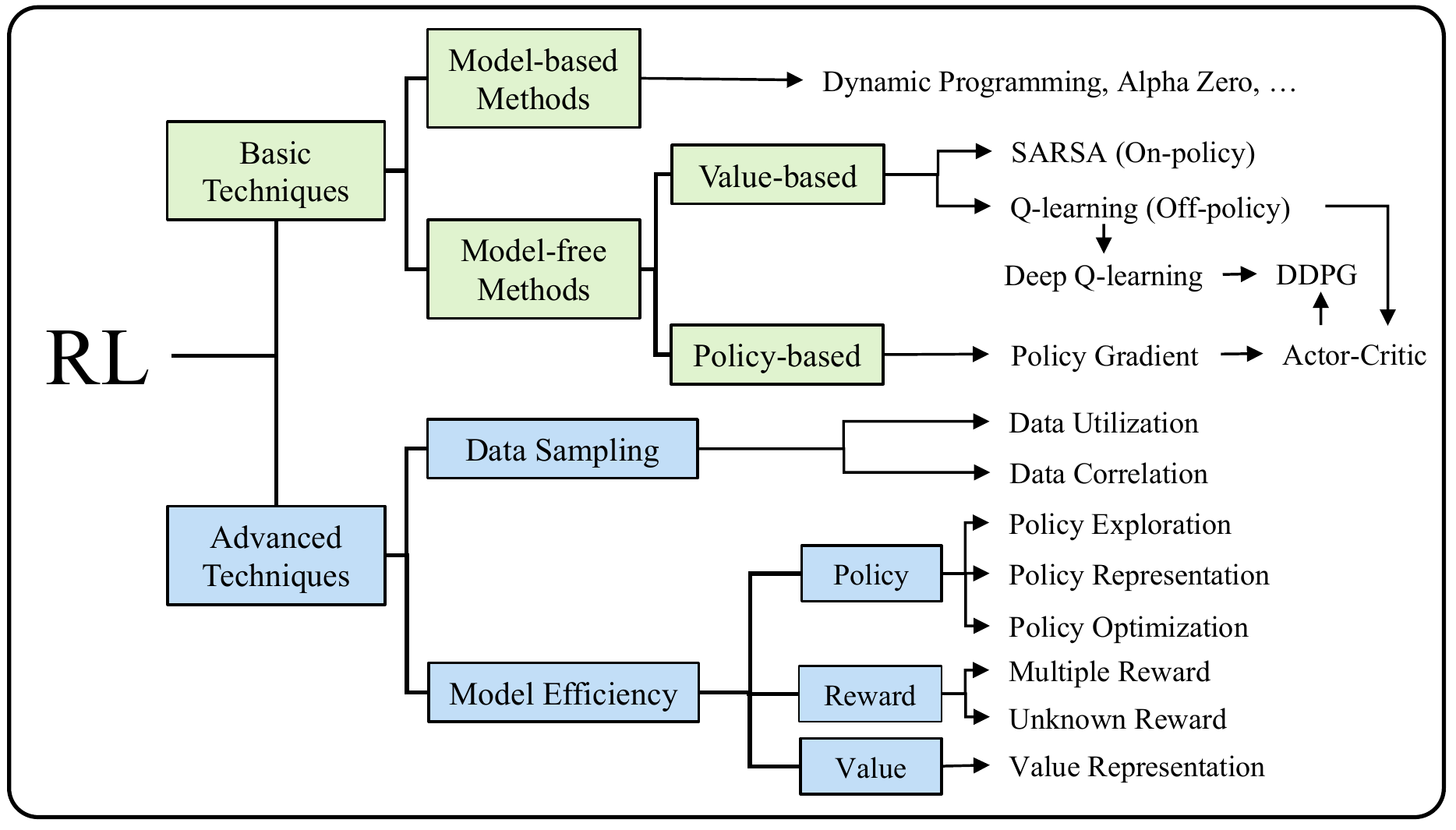}

\caption{Broad categorization of RL techniques.}

\label{fig:techs}

\end{figure*}

By iterating the Bellman equations, we can easily obtain the value functions for a policy, and to compare policies, we define that the policy $\pi$ is better than $\pi'$ if the function $\mathcal{V}$ according to the $\pi$ is no less than the function $\mathcal{V}$ according to the $\pi'$ for all states, that is $\mathcal{V}^{\pi} (s) \geq \mathcal{V}^{\pi'} (s), \forall s$. It has been proven in~\cite{sutton2018reinforcement} that the existence of the optimal policy $\pi^*$ is guaranteed in the MDP problem, where $\mathcal{V}^* (s) =\max_\pi \mathcal{V}^\pi (s)$ and $\mathcal{Q}^* (s) =\max_\pi \mathcal{Q}^\pi (s)$. These two functions are defined as the optimal function $\mathcal{V}$ and the optimal function $\mathcal{Q}$. We can obtain the optimal policy $\pi^*$ by maximizing over the $\mathcal{Q}^*(\pi)$ which can be written as:
\begin{eqnarray}
    \pi^*(a|s) = \arg \max \mathcal{Q}^*(s,a) \label{eq:pi_maxQ}
\end{eqnarray}
To improve the policy, we apply the Bellman optimality equations~\cite{sutton2018reinforcement} to update value functions by taking the action with maximum value instead of trying all possible actions. To facilitate the optimization of the policy, many RL techniques are proposed from different perspectives, and Figure~\ref{fig:techs} provides a diagram outlining the broad categorization of these techniques, illustrating how these techniques can be applied.

\subsection{Basic Techniques}
Based on the representation of MDP elements, basic techniques can be categorized into two classes: \textit{model-based} method and \textit{model-free} method. The main difference is whether the agent has access to model the environment, i.e. whether the agent knows the transition function and the reward function. These two functions are already known in the model-based method where \textit{Dynamic Programming} (DP)\cite{bellman1966dynamic} and \textit{Alpha-Zero} \cite{silver2018general} are the classical methods which have achieved significant results in numerous applications. In these methods, agents are allowed to think ahead and plan future actions with known effects on the environment. Besides, an agent can learn the optimal policy from the planned experience which results in high sample efficiency.

In many RL problems, the reward and the transition function are typically unknown due to the complicated environment and its intricate inherent mechanism. For example, as illustrated in Figure~\ref{pic:QO}, we are unable to obtain the actual latency as the reward in the joint query optimization example. Besides, in the stochastic job scheduling problem~\cite{mao2019variance}, it is also impossible to directly model the transition function because of the randomness of the job arrivals in the practical scenarios. Hence, in these problems, agents usually employ model-free methods that can purely learn the policy from the experience gained during the interaction with the environment. Model-free methods can mainly be classified into two categories, namely the \textit{value-based} method and the \textit{policy-based} method. In the value-based method, the RL algorithm learns the optimal policy by maximizing the value functions. There are two main approaches in estimating the value functions that are \textit{Mento-Carlo} (MC) methods and \textit{Temporal difference} (TD) methods. MC methods calculate the $\mathcal{V}(s)$ by directly applying its definition, that is Equation \ref{eq:V_expectation}. MC methods can directly update the value functions once they get a new trajectory $\tau$ as follows:
\begin{eqnarray}
    \mathcal{V}^\pi(s) \leftarrow \mathcal{V}^\pi(s) + \alpha (\mathcal{G}_{\tau \sim \pi }(\tau | s_0=s) - \mathcal{V}^\pi(s))
\end{eqnarray}
where $\alpha \in [0,1)$ denotes the learning rate which controls the rate of updating the policy with new experiences. However, it has an obvious drawback that a complete trajectory requires the agent to reach a terminal state, while it is not practical in some applications, such as online systems. Different from MC methods, the TD method builds on the recursive relationship of value functions, and hence, can learn from the incomplete trajectory. Mathematically, the update of TD methods can be written as:
\begin{eqnarray}
    \mathcal{V}^\pi(s) \leftarrow \mathcal{V}^\pi(s) + \alpha (R(s,a)+\gamma \mathcal{V}^\pi(s') - \mathcal{V}^\pi(s))
\end{eqnarray}
However, there is bias when estimating the function $\mathcal{V}$ with TD methods because they learn from the recursive relationship. To reduce the bias, TD methods can extend the length of the incomplete trajectories and update the function $\mathcal{V}$ by thinking more steps ahead, which is called $n$-steps TD methods. As $n$ grows to the length of whole trajectories, MC methods can be regarded as a special case of TD methods where function $\mathcal{V}$ is an unbiased estimate. On the other side of the coin, as the length $n$ increases, the variance of the trajectory also increases. In addition to the above consideration, TD-based methods are more efficient and require less storage and computation, thus they are more popular among RL algorithms.

In value-based methods, we can obtain the optimal policy by acting greedily via Equation \ref{eq:pi_maxQ}. The update of the function $\mathcal{Q}$ with TD methods is similar to the update of the function $\mathcal{V}$, and is as follows: $\mathcal{Q}^\pi(s,a) \leftarrow \mathcal{Q}^\pi(s,a) + \alpha (R(s,a)+\gamma \mathcal{Q}^{\pi'}(s',a') - \mathcal{Q}^\pi(s,a))$ where the agent follows the policy $\pi$ to take actions and follows the policy $\pi'$ to maximize the function $\mathcal{Q}$. If the two policies are the same, that is $\pi' = \pi$, we call such RL algorithms the \textit{on-policy} methods where the \textit{SARSA}\cite{rummery1994line} is the representative method. In addition, other policies can also be used in $\pi'$. For example, in \textit{Q-learning}\cite{watkins1992q}, the agent applies the greedy policy and updates the function $\mathcal{Q}$ with the maximum value in its successor. Its update formula can be written as: $\mathcal{Q}^\pi(s,a) \leftarrow \mathcal{Q}^\pi(s,a) + \alpha (R(s,a)+\gamma \max_{a'} \mathcal{Q}^\pi(s',a') - \mathcal{Q}^\pi(s,a))$. Both value-based methods can work well without the model of the environment, and Q-learning directly learns the optimal policy, whilst SARSA learns a near-optimal policy during exploring. Theoretically, Q-learning should converge quicker than SARSA, but its generated samples have a high variance which may suffer from the problems of converging.

In RL, storage and computation costs are very high when there is a huge number of states or actions. To overcome this problem, DRL, as a branch of RL, adopts Deep Neural Network (DNN) to replace tabular representations with neural networks. For function $\mathcal{V}$, DNN takes the state $s$ as input and outputs its state value $\mathcal{V}_\theta (s) \approx \mathcal{V}^\pi (s)$ where the $\theta$ denotes the parameter in the DNN. When comes to function $\mathcal{Q}$, It takes the combination of the state $s$ and the action $a$ as input and outputs the value of the state-action pair $\mathcal{Q}_\theta (s, a) \approx \mathcal{Q}^\pi (s, a)$, As for the neural networks, we can optimize them by applying the techniques that are widely used in deep learning (e.g. gradient descent). \textit{Deep Q-learning network} (DQN) \cite{Mnih2013}, as a representative method in DRL, combines the DNN with Q-learning and its loss function is as follows:
\begin{equation}
    \mathcal{L}_w = \mathbf{E}_\mathcal{D}[(R(s,a) + \gamma \max_{a^*\in \mathcal{A}}\mathcal{Q}_w(s',a^*)-\mathcal{Q}_w(s,a))^2] \label{eq:DeepQ}
\end{equation}
where $\mathcal{D}$ denotes the \textit{experience replay} which accumulates the generated samples and can stabilize the training process.

Policy-based methods are another branch of the model-free RL algorithm that have a clear representation of the policy $\pi(a|s)$, and they can tackle several challenges that are encountered in value-based methods. For example, when the action space is continuous, value-based methods need to discretize the action which could increase the dimensionality of the problem, and memory and computation consumption. Value-based methods learn a deterministic policy that generates the action given a state through an optimal function $\mathcal{Q}$ (i.e. $\pi(s)=a$). However, for policy-based methods, they can learn a stochastic policy (i.e. $\pi_\theta (a_i|s)=p_i, \sum_i p_i=1$) as the optimal policy, where $p_i$ denotes the probability of taking the action $a_i$ given a state $s$, and $\theta$ denotes the parameters where neural networks can be used to approximate the policy. \textit{Policy Gradient} \cite{sutton2000policy} method is one of the main policy-based methods which can tackle the aforementioned challenges. Its goal is to optimize the parameters $\theta$ by using the gradient ascent method, and the target can be denoted in a generalized expression:
\begin{equation}
\nabla_\theta J(\theta)=\mathbf{E}_{\tau \sim \pi_\theta}[R(\tau) \nabla_{\pi_\theta} \log_{\pi_{\theta}}(a|s)]
\end{equation}
The specific proof process can refer to \cite{sutton2018reinforcement}. Sampling via the MC methods, we will get the entire trajectories to improve the policy for the policy-based methods.

After training, the action with higher rewards in expectation will have a higher probability to be chosen and vice versa. As for the continuous action, The optimal policy learned from the Policy Gradient is stochastic which still needs to be sampled to get the action. However, the stochastic policy still requires lots of samples to train the model when the search space is huge. \textit{Deterministic Policy Gradient} (DPG) \cite{silver2014deterministic}, as an extension of the Policy Gradient, overcomes this problem by using a stochastic policy to perform sampling while applying deterministic policy to output the action which demands relatively fewer samples.

Both value-based methods and policy-based methods have their strengths and weaknesses, but they are not contradictory to each other. \textit{Actor-Critic} (AC) method, as the integration of both methods, divides the model into two parts: actor and critic. The actor part selects the action based on the parameterized policy and the critic part concentrates on evaluating the value functions. Different from previous approaches, AC evaluates the advantage function $\mathcal{A}^\pi (s, a) = \mathcal{Q}^\pi (s, a) - \mathcal{V}^\pi (s)$ which reflects the relative advantage of a certain action $a$ to the average value of all actions. The introduction of the value functions also allows AC to update by step through the TD method, and the incorporation of the policy-based methods makes AC be suitable for continuous actions. However, the combination of the two methods also makes the AC method more difficult to converge. Moreover, \textit{Deep Deterministic Policy Gradient} (DDPG) \cite{Lillicrap2016}, as an extension of the AC, absorbs the advanced techniques from the DQN and the DPG which enables DDPG to learn the policy more efficiently.

In all the above-mentioned methods, there always exists a trade-off between exploring the unknown situation and exploiting with learned knowledge. On the one hand, exploiting the learned knowledge can help the model converge quicker, but it always leads the model into a local optimal rather than a globally optimal. On the other hand, exploring unknown situations can find some new and better solutions, but always being in the exploring process causes the model hard to converge. To balance these two processes, researchers have been devoting much energy to finding a good heuristics strategy, such as $\epsilon-greedy$ strategy, Boltzmann exploration (Softmax exploration), upper confidence bound (UCB) algorithm \cite{auer2002using}, Thompson sampling \cite{thompson1933likelihood}, and so on. Here, we consider the $\epsilon-greedy$, a widely used exploration strategy, as an example. $\epsilon-greedy$ typically selects the action with the maximal Q value to exploit the learned experience while occasionally selecting an action evenly at random to explore unknown cases. $\epsilon-greedy$ exploration strategy with $m$ actions can be denoted as follow:
\begin{equation}
\pi(a|s)=
\left\{
             \begin{array}{ll}
             \epsilon/m + (1-\epsilon) & a^* = arg\max_{a\in \mathcal{A}}\mathcal{Q}(s,a), \\
             \epsilon/m & a \not= a^*.
             \end{array} \label{e-greedy}
\right.
\end{equation}
$\epsilon \in [0,1)$ is an exploration factor. The agent is more likely to select the action at random when the $\epsilon$ is closer to 1, and the $\epsilon$ will be continuously reduced during the training process.

\subsection{Advanced Techniques}

This section mainly discusses some advanced techniques in RL which focus on efficiently using the limited samples and building sophisticated model structures for better representation and optimization. According to the different improvements, they can be broadly classified into two parts: data sampling and model efficiency.

\subsubsection{Data Sampling}
Data sampling is one of the most important concerns in training the DRL in data processing and analytics. In most applications, the sample generation process costs a great amount of time and computation resources. For example, a sample may refer to an execution run for workload and repartitioning for the database, which can take about 40 minutes\cite{hilprecht2020learning}. Hence, to train the model with limited samples, we need to increase data utilization and reduce data correlation.

\textbf{Data utilization}:
Most DRL algorithms train the optimal policy and sample data at the same time. Instead of dropping samples after being trained, \textit{experience replay}\cite{lin1992self} accumulates the samples in a big table where samples are randomly selected during the learning phase. With this mechanism, samples will have a higher utilization rate and a lower variance, and hence, it can stabilize the training process and accelerate the training convergence. Samples after several iterations may differ from the current policy, and hence, \textit{Growing-batch} \cite{lange2012batch} can continuously refresh the table and replace these outdated samples. In addition, samples that are far away from the current policy should be paid more attention and \textit{Prioritized Experience Replay}\cite{schaul2016prioritized} uses TD error as the priority to measure the sample importance, and hence, focus more on learning the samples with high errors. In a nutshell, with the experience replay, DRL cannot only stable the learning phase but also efficiently optimize the policy with fewer samples.

\textbf{Data correlation}:
Strong correlation of training data is another concern that may lead the agent to learn a sub-optimal solution instead of the globally optimal one. Apart from the experience replay, the mechanism of the distributed environments is another research direction to alleviate this problem. For example, the \textit{asynchronous advantage actor-critic} (A3C) \cite{Mnih2016} and \textit{Distributed PPO} (DPPO) \cite{Heess2017} apply multi-threads to build multiple individual environments where multiple agents take actions in parallel, and the update is calculated periodically and separately which can accelerate the sampling process and reduce the data correlation.

\subsubsection{Model Efficiency}

\label{Model Efficiency}
RL model with better efficiency is the major driving force of the development of RL, and there are many researchers improving it from three major aspects, namely policy, reward function, and value function.

\textbf{Policy}:
The policy-related techniques focus on stably and effectively learning a comprehensive policy, and the advanced techniques to efficiently learn the policy can be classified into three parts in detail, which are policy exploration, policy representation, and policy optimization.

a) \textit{Policy exploration}: Its target is to explore as many actions as possible during the training process in case the policy will be trapped into the local optimal. For example, \textit{entropy regularisation} \cite{Mnih2016} adds the entropy of the actions' probabilities into the loss item which can sufficiently explore the actions. Besides, adding noise to the action is another research direction to increase the randomness into policy exploration. The DDPG applies an Ornstein–Uhlenbeck process \cite{uhlenbeck1930theory} to generate temporal noise $\mathcal{N}$ which are directly injected into policy.
\textit{Noisy-Net} \cite{Fortunato2018} incorporates the noise into the parameters of neural networks which is easy to implement, and it shows a better performance than the $\epsilon-greedy$ and entropy regularisation methods.
Further, Plappert et al. \cite{Plappert2018} investigate an effective way to combine the parameter space noise to enrich the exploratory behaviors which can benefit both on-policy methods and off-policy methods.

b) \textit{Policy representation}: The states in some RL problems are in a huge dimension which causes challenges during the training. To approximate a better policy, a branch of DRL models improve the policy representation by absorbing convolutional neural networks (CNN) into DQN to analyze the data, such as \textit{Dueling DQN} \cite{Wang2016}, \textit{DRQN} \cite{Hausknecht2015}, and so on. In addition, DRQN also incorporates the LSTM structure to increase the capacity of the policy which is able to capture the temporal information, such as speed, direction.

c) \textit{Policy optimization}: The update of the value functions following the Equation \ref{eq:bellman_v} and \ref{eq:bellman_q} tends to overestimate the value functions and introduce a bias because they learn estimates from the estimates. Mnih et al.\cite{mnih2015human} separate the two estimation process by using two same Q-networks which can reduce the correlation of two estimation processes and hence, stabilize the course of training. However, the action with the maximum Q-value may differ between two Q-networks which will be hard for convergence. and \textit{Double DQN} (DDQN) \cite{hasselt2015doubledqn} alleviate the issue by disaggregating the step of selecting the action and calculating the max Q-value.

When we apply the policy-based RL methods, the learning rate of the policy plays an essential role in achieving superior performance. A higher learning rate can always maximize the improvement on a policy by step, but it also causes the instability of the learning phase. Hence, The \textit{Trust Region Policy Optimization} (TRPO) \cite{Schulman2015} builds constraints on the old policy and new policy via KL divergence to control the change of the policy in an acceptable range. With this constraint, TRPO can iteratively optimize the policy via a surrogate objective function which can monotonically improve policies. However, the design of the KL constraint makes it hard to be trained, and \textit{Proximal Policy Optimization} (PPO) \cite{Schulman2017} simplifies the constraint through two ways: adding it into the objective function, designing a clipping function to control the update rate. Empirically, PPO methods are much simpler to implement and are able to perform at least as well as TRPO.

\textbf{Reward}:
Reward function as one of the key components in the MDP plays an essential role in the RL. In some specific problems, the agent has to achieve multiple goals which may have some relationships. For example, the robot can only get out through the door only if it has already found the key. To tackle this challenge, \textit{Hierarchical DQN} \cite{Kulkarni2016} proposes two levels of hierarchical RL (HRL) models to repeatedly select a new goal and achieve the chosen goal. However, there is a limitation that the goal needs to be manually predefined which may be unknown or unmeasurable in some environments, such as the market and the effect of a drug. To overcome it, \textit{Inverse RL} (IRL) \cite{ng2000algorithms} learns the rewards function from the given experts' demonstrations (i.e. the handcraft trajectories), but the agent in IRL can only prioritize the entire trajectories over others. It will cause a shift when the agent comes to a state that never appears before, and \textit{Generative Adversarial Imitation Learning} (GAIL) \cite{ho2016generative}, as an imitation learning algorithm, applies adversarial training methods to generate fake samples and is able to learn the expert's policy explicitly and directly.

\textbf{Value}:
As we have mentioned earlier, the tabular representation of the value functions has several limitations which can be alleviated via DRL. Different from directly taking the state-action pair as the input to calculate the Q-function, \textit{Dueling DQN} \cite{Wang2016} estimates its value by approximating two separate parts that are the state-values and the advantage values, and hence, can distinguish whether the value is brought by the state or the action.

The aforementioned advanced algorithms and techniques improve and enhance the DRL from different perspectives, which makes DRL-based algorithms be a promising way to improve data processing and analytics. We observe that problems with the following characteristics may be amenable to DRL-based optimization. First, problems are incredibly complex and difficult. The system and application involve a complicated operational environment (e.g., large-scale, high-dimensional states) and internal implementation mechanisms, which is hard to construct a white-box model accurately. DRL can process complex data and learn from experience generated from interacting, which is naturally suitable for data processing and analytics where many kinds of data exist and are processed frequently. Second, the optimization objectives can be represented and calculated easily as the reward because the RL agent improves itself towards maximizing the rewards and rewards could be computed a lot of times during training. Third, the environment can be well described as MDP. DRL has been shown to solve MDP with theoretical guarantees and empirical results. Thus, problems involving sequential decision making such as planning, scheduling, structure generation (e.g., tree, graph), and searching could be expressed as MDP and a good fit for DRL. Fourth, collecting required labels of data massively is hard. Compared to supervised learning, DRL can utilize data efficiently to gain good performance.

\section{
Data System Optimizations}

DRL learns knowledge about the system by interacting with it and optimizes the system. In this section, we focus on several fundamental aspects with regards to system optimization in data processing and analytics including data organization, scheduling, tuning, indexing, query optimization, and cache management. We discuss how each problem is formulated in MDP by defining three key elements (action, state, and reward) in the system and solved by DRL. Generally, the states are defined by some key characteristics of the system. The actions are possible decisions (e.g., system configuration), that affect the system performance and the reward is calculated based on the performance metrics (e.g. throughput, latency). Table 1 presents a summary of representative works and the estimated dimension ranges on the state and action space of each work are added as signals on the DRL training difficulty. As a comparison, OpenAI Five\cite{openai}, a Dota-playing AI, observes the state as 20,000 numbers representing useful game information and about 1,000 valid actions (like ordering a hero to move to a location) for per hero. Dota is a real-time strategy game between two teams of five players where each player controls a character called a “hero”.

\subsection{Data Organization}

\subsubsection{Data Partitioning}
Effective data partitioning strategy is essential to accelerate data processing and analytics by skipping irrelevant data for a given query. It is challenging as many factors need to be considered, including the workload and data characteristics, hardware profiles, and system implementation.

In data analytics systems, data is split into blocks in main memory or secondary storage, which are accessed by relevant queries. A query may fetch many blocks redundantly and, therefore, an effective block layout avoids reading unnecessary data and reduces the number of block accesses, thereby improving the system performance. Yang et al.\cite{yang2020qd} propose a framework called the qd-tree that partitions data into blocks using DRL over the analytical workload. The qd-tree resembles the classic k-d tree and describes the partition of multi-dimensional data space where each internal node splits data using a particular predicate and represents a subspace. The data in the leaf node is assigned to the same block. In the MDP, each state is a node representing the subspace of the whole data and featured as the concatenation of range and category predicates. After the agent takes an action to generate two child nodes, two new states will be produced and explored later. The available action set is the predicates parsed from workload queries. The reward is computed by the normalized number of skipped blocks over all queries. They do not execute queries and a sampling technique is used to estimate the reward efficiently. The formulation of using DRL to learn a tree is similar to NeuroCuts\cite{liang2019neural} that learns a tree for packet classification. However, the qd-tree may not support a complex workload containing user-defined functions (UDFs) queries.

Horizontal partitioning in the database chooses attributes of large tables and splits them across multiple machines to improve the performance of analytical workloads. The design relies on either the experience of database administrators (DBAs) or cost models that are often inaccurate\cite{leis2015good} to predict the runtime for different partitions. Data collection is too challenging and costly to train the accurate supervised learning model in the cloud environment. Hilprecht et al.\cite{hilprecht2020learning} learn to partition using DRL on analytical workloads in cloud databases, on the fact that DRL is able to efficiently navigate the partition search and requires less training data. In the MDP, the state consists of two parts. The database part encodes whether a table is replicated, an attribute is used for partitioning, and which tables are co-partitioned. The workload part incorporates normalized frequencies of representative queries. Supported actions are: partitioning a table using an attribute, replicating a table, and changing tables co-partition. The reward is the negative of the runtime for the workload. One challenge is that the cost of database partitioning is high during training. To alleviate the problem, the agent is trained in the simulation environment and is further refined in the real environment by estimating the rewards using sampling. One limitation is that it may not support new queries well because only the frequency features of queries are considered. Durand et al. in \cite{durand2018gridformation, durand2019automated} utilize DRL to improve vertical partitioning that optimizes the physical table layout. They show that the DQN algorithm can easily work for a single workload with one table but is hard to generalize to random workloads.

For UDFs analytics workloads on unstructured data, partitioning is more challenging where UDFs could express complex computations and functional dependency is unavailable in the unstructured data. Zou et al.\cite{zou2020lachesis} propose the Lachesis system to provide automatic partitioning for non-relational data analytics. Lachesis translates UDFs to graph-based intermediate representations (IR) and identifies partition candidates based on the subgraph of IR as a two-terminal graph. Lachesis adopts DRL to learn to choose the optimal candidate. The state incorporates features for each partition extracted from historical workflows: frequency, the execution interval, time of the most recent run, complexity, selectivity, key distribution, number, and size of co-partition. In addition, the state also incorporates other features such as hardware configurations. The action is to select one partition candidate. The reward is the throughput speedup compared to the average throughput of the historical executions of applications. To reduce the training time, the reward is derived from historical latency statistics without partitioning the data when running the applications. One limitation is that Lachesis largely depends on historical statistics to design the state and calculate the reward, which could lead to poor performance when the statistics are inadequate.

\subsubsection{Data Compression}
Data compression is widely employed to save storage space. The effectiveness of a compression scheme however relies on the data types and patterns. In time-series data, the pattern can change over time and a fixed compression scheme may not work well for the entire duration. Yu et al.\cite{yu2020two} propose a two-level compression framework, where a scheme space is constructed by extracting global features at the top level and a compression schema is selected for each point at the bottom level. The proposed AMMMO framework incorporates compression primitives and the control parameters, which define the compression scheme space. Due to the fact that the enumeration is computationally infeasible, the framework proposes to adopt DRL to find the compression scheme. The agent takes a block that consists of 32 data points with the compressed header and data segment, timestamps, and metrics value as the state. The action is to select a scheme from compression scheme space and then the compression ratio is computed as the reward. The limitation is that the method may not work for other data types like images and videos.

\begin{table*}[t]

\footnotesize 
\centering

\caption{Representative Works using DRL for Data System Optimizations. D(X) denotes the approximate dimension of X space. }

\begin{tabular}{|p{1.4cm}|p{2.7cm}|p{1.6cm}|p{1.3cm}|p{1.5cm}|p{3.2cm}|p{1.1cm}|}

\hline\hline

\textbf{Domain} & \textbf{Work} & \textbf{Algorithm} & \textbf{D(State)}& \textbf{D(Action)} & \textbf{DRL-based Approach}&\textbf{Open Source}\\

\hline
Data organization &Analytical system data partition\cite{yang2020qd} & PPO & 10 - 100 & 100 - 1000 & Exploit workload patterns and Generate the tree &NO \\
\cline{2-7}
&Database horizontal partition~\cite{hilprecht2020learning}&DQN&100&10 & Navigate the partition search efficiently&NO \\
\cline{2-7}
&UDF-centric workload data partition~\cite{zou2020lachesis}&A3C&10& 1-10 & Exploit the features of partition and search&YES \\
\cline{2-7}
&Time series data compression~\cite{yu2020two}&PG&100& 10& Search parameters interactively& NO \\

\hline
Scheduling &Distributed job processing~\cite{mao2019learning}&PG&100& 10 & Exploit the job dependencies and learn schedule decision&YES\\
\cline{2-7}
&Distributed stream data~\cite{li2018model}&DDPG&100& 10-100 & Learn schedule decision&NO \\

\hline
Tuning &Database configuration~\cite{zhang2019end}~\cite{li2019qtune}&DDPG&100 & 10 & Search configuration parameters interactively&YES \\

\hline
Index &Index Selection~\cite{sharma2018case}&CEM&100& 10& Search the index interactively&NO\\
\cline{2-7}
&R-tree construction~\cite{gu2021rlr}&DQN& 10-100& 10 & Learn to generate the tree&NO \\

\hline
Query Optimization &Join order selection~\cite{marcus2018deep, krishnan2018learning, yu2020reinforcement, heitz2019join}&PG, DQN, ...&10-100&1-10&Learn to decide the join order&Only~\cite{heitz2019join} \\

\hline
Cache Management &View Materialization~\cite{yuan2020automatic}&DQN&100 & 10 & Model the problem as IIP and solve&NO \\

\hline

\hline

\end{tabular}

\label{tab:1}

\end{table*}

\subsection{Scheduling}
Scheduling is a critical component in data processing and analytics systems to ensure that resources are well utilized. Job scheduling in a distributed computing cluster faces many challenging factors such as workload (e.g., job dependencies, sizes, priority), data locality, and hardware characteristics. Existing algorithms using general heuristics such as shortest-job-first do not utilize these factors well and fail to yield top performance. To this end, Mao et al.\cite{mao2019learning} propose Decima to learn to schedule jobs with dependent stages using DRL for data processing clusters and improve the job completion time. In the data processing systems such as Hive\cite{thusoo2009hive}, Pig\cite{olston2008pig}, Spark-SQL\cite{armbrust2015spark}, jobs could have up to hundreds of stages and many stages run in parallel, which are represented as directed acyclic graphs (DAGs) where the nodes are the execution stages and each edge represents the dependency. To handle parallelism and dependencies in job DAGs, Decima first applies graph neural network (GNN) to extract features as the state instead of manually designing them while achieving scalability. Three types of feature embeddings are generated. Node embedding captures information about the node and its children including the number of remaining tasks, busy and available executors, duration, and locality of executors. Job embedding aggregates all node embeddings in the job and cluster embedding combines job embeddings. To balance possible large action space and long action sequences, The action determines the job stage to be scheduled next and the parallelism limit of executors. The reward is based on the average job completion time. To train effectively in a job streaming environment, Decima gradually increases the length of training jobs to conduct curriculum learning\cite{bengio2009curriculum}. The variance reduction technique\cite{mao2019variance} is applied to handle stochastic job arrivals for robustness. However, we note that Decima is non-preemptive and does not re-schedule for higher priority jobs.

In distributed stream data processing, streams of continuous data are processed at scale in a real-time manner. The scheduling algorithm assigns workers to process data where each worker uses many threads to process data tuples and aims to minimize average data tuple processing time. Li et al.\cite{li2018model} design a scheduling algorithm using DRL for distributed stream data processing, which learns to assign tuples to work threads. The state consists of the scheduling plan (e.g., the current assignment of workers) and the workload information (e.g., tuple arrival rate). The action is to assign threads to machines. The reward is the negative tuple processing time on average. The work shows that DQN does not work well because the action space is large and applies DDPG to train the \textit{actor-critic} based agent instead. To find a good action, the proposed method looks for k nearest neighbors of the action that the \textit{actor} network outputs and selects the neighbor with the highest value that the \textit{critic} network outputs. The algorithm is implemented on Apache Storm and evaluated with representative applications: log stream processing, continuous queries, and word count.

Many works have been recently proposed to improve scheduling using DRL\cite{zhang2020buffer, kraska2019sagedb}. Query scheduling determines the execution order of queries, which has a great influence on query performance and resource utilization in the database system. SmartQueue\cite{zhang2020buffer} improves query scheduling by leveraging overlapping data access among queries and learns to improve cache hits using DRL. In addition, Tim et al.\cite{kraska2019sagedb} design a scheduling system in SageDB using RL techniques. Other works using RL for scheduling include Bayesian RL for scheduling in heterogeneous clusters\cite{banerjee2020inductive}, operation scheduling in devices\cite{gao2018spotlight}, application container scheduling in clusters\cite{10.5555/3433701.3433791}, etc.

\subsection{Tuning}
Tuning the configuration of data processing and analytic systems plays a key role to improve system performance. The task is challenging because up to hundreds of parameters and complex relations between them could exist. Furthermore, other factors such as hardware and workload also impact the performance. Existing works often employ search-based or supervised learning methods. The former takes much time to get an acceptable configuration and the latter such as OtterTune\cite{van2017automatic} needs large high-quality data that is non-trivial to obtain in practice. Zhang et al.\cite{zhang2019end} design a cloud database tuning system CDBTune using DRL to find the best parameter in high-dimensional configuration space. The CDBTune formulates MDP as follows. The state is represented by the internal metrics (e.g., buffer size, pages read). The action is to increase or decrease the knob values. The reward is the performance difference between two states, which is calculated using throughput and latency. CDBTune takes several hours on offline training in simulation and online training in the real environment. Compared to OtterTune, CDBTune eases the burden of collecting large training data sets. In the experiments, CDBTune is shown to outperform DBA experts and OtterTune and improve tuning efficiency under 6 different workloads on four databases. One limitation of the approach is that the workload information is ignored and thus it may not perform well when the query workload is changed.

To address the issue, Li et al.\cite{li2019qtune} propose QTune that considers query information to tune the database using DRL. First, Qtune extracts features from SQL query including types (e.g., insert, delete), tables, and operation (e.g., scan, hash join) costs estimated by the database engine. The columns attributes and operations like selection conditions in the query are ignored. Subsequently, Qtune trains a DNN model to predict the difference of statistics (e.g., updated tuples, the number of committed transactions) in the state after executing the queries in the workload and updates the state using it. The action and reward design are similar to CDBTune. Additionally, QTune supports three levels of tuning granularity for balancing throughput and latency. For query-level, QTune inputs query vector and tries to find good knobs for each query. For workload-level, vectors for all queries are merged and used. For cluster-level, QTune employs a clustering method based on deep learning to classify queries and merge queries into clusters. One drawback of QTune is that the query featurization could lose key information such as query attributes (i.e., columns) and hurt the performance especially when the cost estimation is inaccurate. The prediction model for state changes is trained alone and needs accurate training data. An end-to-end training framework is therefore essential and a good direction to undertake.

\subsection{Indexing}

\subsubsection{Database Index Selection}
Database index selection considers which attributes to create an index to maximize query performance. Sharma et al.\cite{sharma2018case} show how DRL can be used to recommend an index based on a given workload. The state encodes selectivity values for workload queries and columns in the database schema and current column indexes. The action is to create an index on a column. The reward is the improvement compared to the baseline without indexes. The experiments show that the approach can perform as well or better as having indexes on all columns. Sadri et al.\cite{sadri2020drlindex} utilize DRL to select the index for a cluster database where both query processing and load balancing are considered. Welborn et al.\cite{welborn2019learning} optimize the action space design by introducing task-specific knowledge for index selection tasks in the database. However, these works only consider the situation where single-column indexes are built. Lan et al.\cite{lan2020index} propose both single-attribute and multi-attribute indexes selection using DRL. Five rules are proposed to reduce the action and state space, which help the agent learn effective strategy easier. The method uses what-if caller\cite{chaudhuri1998autoadmin} to get the cost of queries under specific index configurations without building indexes physically. These works conduct basic experiments with small and simple datasets. Extensive and large-scale experiments using real datasets are therefore needed to benchmark these methods to ensure that they can scale well.

\subsubsection{Index Structure Construction}
The learned index is proposed recently as an alternative index to replace the B$^+$-Tree and bloom filter by viewing indexes as models and using deep learning models to act as indexes\cite{kraska2018case}. DRL can enhance the traditional indexes instead of replacing them.

Hierarchical structures such as the B$^+$-tree and R-tree are important indexing mechanisms to locate data of interest efficiently without scanning a large portion of the database. Compared to the single dimensional counterpart, the R-tree is more complex to optimize due to bounding box efficiency and multi-path traversals. Earlier conventional approaches use heuristics to determine these two operations (i.e. choosing the insertion subtree and splitting an overflowing node) during the construction of the R-tree\cite{ooi1993indexing}. Gu et al.\cite{gu2021rlr} propose to use DRL to replace heuristics to construct the R-tree and propose the RLR-tree. The approach models two operations ChooseSubtree and Split as two MDPs respectively and combines them to generate an R-Tree. For ChooseSubtree, the state is represented as the concatenation of the four features (i.e., area, perimeter, overlap, occupancy rate) of each selected child node. More features are evaluated but do not improve the performance in the reported experiments. The action is to select a node to insert from top-k child nodes in terms of the increase of area. The reward is the performance improvement from the RLR-tree. For Split MDP, the state is the areas and perimeters of the two nodes created by all top-k splits in the ascending order of total area. The action is to choose one split rule from k rules and the reward is similar to that of ChooseSubtree. The two agents are trained alternately. As expected, the optimizations render the RLR-tree improved performance in range and KNN queries.

Graphs can be used as effective indexes to accelerate nearest neighbors search\cite{malkov2018efficient, dong2011efficient}. Existing graph construction methods generally propose different rules to generate graphs, which cannot provide adaptivity for different workloads\cite{wang2021comprehensive}. Baranchuk et al.\cite{baranchuk2019similarity} employ DRL to optimize the graph for nearest neighbors search. The approach learns the probabilities of edges in the graph and tries to maximize the search efficiency. It considers the initial graph and the search algorithm as the state. The action is to keep an edge or not. The reward is the performance for search. It chooses the TRPO\cite{Schulman2015} algorithm to train. The reported experimental results show that the agent can refine state-of-the-art graphs and achieve better performance. However, this approach does not learn to explore and add new edges to the initial graph that may affect the performance.

Searching and constructing a new index structure is another line of interesting research~\cite{idreos2019learning}. Inspired by Neural Architecture Search (NAS)\cite{zoph2016neural}, Wu et al.\cite{wu2019progressive} propose an RNN-based neural index search (NIS) framework that employs DRL to search the index structures and parameters given the workload. NIS can generate tree-like index structures layer by layer via formalizing abstract ordered blocks and unordered blocks, which can provide a well-designed search space. The keys in the ordered block are sorted in ascending order, and the skip list or B$^+$-Tree can be used. The keys in the unordered block are partitioned using customized functions and the hash bucket can be used. Overall, the whole learning process is similar to that of NAS.

\subsection{Query Optimization}
Query optimization aims to find the most efficient way to execute queries in database management systems. There are many different plans to access the query data that can have a large processing time variance from seconds to hours. The performance of a query plan is determined mostly by the table join orders. Traditionally, query optimizers use certain heuristics combined with dynamic programming to enumerate possible efficient execution plans and evaluate them using cost models that could produce large errors\cite{leis2015good}. Marcus et al.\cite{marcus2018deep} propose Rejoin that applies DRL to learn to select better join orders utilizing past experience. The state encodes join tree structure and join predicates. The action is to combine two subtrees, where each subtree represents an input relation to join. The reward is assigned based on the cost model in the optimizer. The experiments show that ReJOIN can match or outperform the optimizer in PostgreSQL. Compared to ReJoin, DQ\cite{krishnan2018learning} presents an extensible featurization scheme for state representation and improves the training efficiency using the DQN\cite{Mnih2013} algorithm. Heitz et al.\cite{heitz2019join} compare different RL algorithms including DQN\cite{Mnih2013}, DDQN\cite{Lillicrap2016}, and PPO\cite{Schulman2017} for join order optimization and use a symmetric matrix to represent the state instead of vector. Yu et al.\cite{yu2020reinforcement} introduce a graph neural network (GNN) with DRL for join order selection that replaces fixed-length hand-tuned vector in Rejoin\cite{marcus2018deep} and DQ\cite{krishnan2018learning} with learned scalable GNN representation and better captures and distinguishes the join tree structure information. These works mainly differ in encoding what information and how to encode them.

Instead of learning from past query executions, Trummer et al.\cite{trummer2019skinnerdb} propose SkinnerDB to learn from the current query execution status to optimize the remaining execution of a query using RL. Specifically, SkinnerDB breaks the query execution into many small time intervals (e.g., tens to thousands of slices per second) and processes the query adaptively. At the beginning of each time interval, the RL agent chooses the join order and measures the execution progress. SkinnerDB adopts a similar adaptive query processing strategy in Eddies\cite{tzoumas2008reinforcement} and uses the UCT algorithm\cite{kocsis2006bandit}, which provides formal guarantees that the difference is bounded between the rewards obtained by the agent and those by optimal choices. The reward is calculated by the progress for the current interval. A tailored execution engine is designed to fully exploit the learning strategy with tuple representations and specialized multi-way join algorithms. SkinnerDB offers several advantages. First, it is inherently robust to query distribution changes because its execution only depends on the current query. Second, it relies on less assumption and information (e.g., cardinality models) than traditional optimizers and thus is more suitable for the complicated environment where cardinality is hard to estimate. Third, it predicts the optimal join order based on real performance. However, it may introduce overhead caused by join order switching.

Learning-based methods that have been proposed to replace traditional query optimizers often incur a great deal of training overhead because they have to learn from scratch. To mitigate the problem, Bao \cite{marcus2021bao} (the Bandit optimizer)) is designed to take advantage of the existing query optimizers. Specifically, Bao learns to choose the best plan from the query plan candidates provided by available optimizers by passing different flags or hints to them. Bao transforms query plan trees into vectors and adopts a tree convolutional neural network to identify patterns in the tree. Then it formulates the choosing task as a contextual multi-armed bandit problem and uses Thompson sampling\cite{thompson1933likelihood} to solve it. Bao is a hybrid solution for query optimization. It achieves good training time and is robust to changes in workload~\cite{marcus2021bao}.
\begin{table}[t]
\footnotesize
\centering

\caption{Methods of query optimization.}

\begin{center}

\begin{tabular}{ |p{2.5cm}|p{2.6cm}|p{1.5cm}|p{1.8cm}|}

\hline
Method&Techniques&Training&Workload Adaptivity
\\

\hline
Rejoin\cite{marcus2018deep}, DQ\cite{krishnan2018learning}&learn from execution experience & High & Low \\
\hline
SkinnerDB~\cite{trummer2019skinnerdb}&learn from current execution status & Medium & Medium
\\

\hline
Bao\cite{marcus2021bao}&learn to choose existing optimizers & Low & High \\
\hline

\end{tabular}

\end{center}

\end{table}

\subsection{Cache Management}

\subsubsection{View Materialization}
View materialization is the process of deciding which view, i.e., results of query or subquery, to cache. In database systems, a view is represented as a table and other queries could be accelerated by reading this table instead of accessing the original tables. There is an overhead of materializing and maintaining the view when the original table is updated. Existing methods are based on heuristics, which either rely on simple Least-Recently-Used rule or cost-model based approaches\cite{perez2014history}. The performance of these approaches is limited because feedback from the historical performance of view materialization is not incorporated. Liang et al.\cite{liang2019opportunistic} implement Deep Q-Materialization (DQM) system that leverages DRL to improve the view materialization process in the OLAP system. First, DQM analyzes SQL queries to find candidate views for the current query. Second, it trains a DRL agent to select from the set of candidates. Third, it uses an eviction policy to delete the materialized views. In the MDP, the state encodes view state and workload information. The action is to create the view or do nothing. The reward is calculated by the query time improvement minus amortized creation cost. Additionally, the eviction policy is based on credit and it evicts the materialized view with the lowest score.

Yuan et al.\cite{yuan2020automatic} present a different way that use DRL to automate view generation and select the most beneficial subqueries to materialize. First, the approach uses a DNN to estimate the benefits of a materialized view where features from tables, queries, and view plans are extracted. Then the approach models selection as an Integer Linear Programming (IIP) problem and introduce an iterative optimization method to figure it out. However, the method cannot guarantee convergence. To address the issue, the problem is formulated as the MDP. The state encodes the subqueries that are selected to materialize and status if queries use these materialized views. The action is to choose the subquery to materialize or not. The reward is the difference between benefit changes of two states. Both cost estimation and view selection models are trained offline using the actual cost of queries and benefits. Then the cost estimation model is used for the online recommendation for view materialization. Performance study shows its good performance; However, it lacks a comparison with DQM.

\subsubsection{Storage}
Cache management impacts the performance of computer systems with hierarchical hardware structures directly. Generally, a caching policy considers which objects to cache, to evict when the cache is full to maximize the object hit rate in the cache. In many systems, the optimal caching policy depends on workload characteristics. Phoebe\cite{wu2020phoebe} is the RL-based framework for cache management for storage models. The state encodes the information from a preceding fixed-length sequence of accesses where for each access, nine features are extracted including data block address, data block address delta, frequency, reuse distance, penultimate reuse distance, average reuse distance, frequency in the sliding window, the number of cache misses, and a priority value. The action is to set a priority value ranging within $[-1, 1]$ to the data. The reward is computed from if the cache is hit or missed and values are 1 and -1 respectively. It applies the DDPG algorithm to train the agent. Periodical training is employed to amortize training costs in online training. In network systems, one issue is that the reward delay is very long in systems with a large cache, i.e., CDN cache can host up to millions of objects. Wang et al.\cite{wanglearning} propose a subsampling technique by hashing the objects to mitigate the issue when applying RL on caching systems.

\section{Data Analytics Applications}

\begin{table*}[htp]

\footnotesize
\centering

\caption{Representative works for RL applications. D(X) denotes the approximate dimension of X space.}

\begin{tabular}{|p{1.6cm}|p{3.0cm}|p{1.5cm}|p{1.3cm}|p{1.5cm}|p{3.5cm}|}

\hline\hline

\textbf{Domain} & \textbf{Work} & \textbf{Algorithm} & \textbf{D(State)} & \textbf{D(Action)} & \textbf{DRL-based Approach}\\

\hline
Data processing &Entity matching\cite{clark-manning-2016-deep, DBLP.abs-1902-00330} & PG & 100 - 1000 & 100 - 1000 & Select target entity from the candidate entities \\
\cline{2-6}
application &Database interaction with natural language~\cite{zhong2017seq2sql, dong-lapata-2016-language}&PG& 100 - 1000 &100 - 1000 & Learn to generate the query \\
\cline{2-6}
&Feature engineering~\cite{10.1145/3292500.3330868}&DQN&100& 1-10 & Select features and model feature correlations in states \\
\cline{2-6}
&Exploratory data analysis~\cite{10.1145/3318464.3389779}&A3C&10-100 & 100000 & Learn to query a dataset for key characteristics \\
\cline{2-6}
&Abnormal detection~\cite{10.1145/3292500.3330932}&IRL&1-10 & 1-10& Learn the reward function for normal sequences \\
\cline{2-6}
&AutoML pipeline generation~\cite{10.1145/3394486.3403261}&DQN& 10 & 100 & Learn to select modules of a pipeline \\

\hline
Healthcare &Treatment recommendation~\cite{10.1145/3219819.3219961}&DDPG&10 & 100-1000 & Select treatment from candidate treatments\\
\cline{2-6}
&Diagnostic inference~\cite{Ling2017DiagnosticIV}&DQN&100-1000& 1-10 & Learn diagnostic decision \\
\cline{2-6}
&Hospital resource allocation~\cite{pmlr-v119-el-bouri20a}&DDPG&100& 1000-10000 & Learn resource scheduling \\

\hline
Fintech &Portfolio optimization~\cite{dixon2020glearner}&Q-Learning&100 & 100 & Select the portfolio weights for stocks \\
\cline{2-6}
&Trading~\cite{YANG2018388, articleYang}&IRL& 1-10 & 10 & Learn the reward function of trading behaviors \\
\cline{2-6}
&Fraud detection~\cite{articleYang}&IRL&100& 10-100 & Learn the reward function of trading behaviors \\

\hline
E- &Online advertising~\cite{10.1145/3219819.3219918}&DQN&1-10& 1-10& Learn to schedule the advertisements\\
\cline{2-6}
Commerce &Online recommendation~\cite{10.1145/3219819.3220122}&DQN& 100 & 10000 & Learn to schedule recommendations \\
\cline{2-6}
&Search results aggregation~\cite{10.1145/3308558.3313455}&DQN& 10-100& 10-100 & Learn to schedule search results \\

\hline
Others &User profiling~\cite{10.1145/3394486.3403128}&DQN&100-1000&1000-10000&Select users' next activities by modeling spatial semantics \\
\cline{2-6}
&Spammer detection~\cite{10.1145/3394486.3403135}&PG& 100& 100 & Search for the detector by interacting with spammers \\
\cline{2-6}
&Transportation~\cite{10.1145/3292500.3330933}&PG& 1000-10000& 1000 & Learn to schedule transportation \\

\hline

\hline

\end{tabular}

\label{tab:3}

\end{table*}

In this section, we shall discuss DRL applications from the perspective of data processing and data analytics. These two categories of DRL applications form indispensable parts of a pipeline, in which data processing provides a better basis for data analytics. In addition, these two categories share some overlapping topics, making these topics mutually motivating and stimulating. We have summarized the technical comparisons of different applications in Table \ref{tab:3}. We shall first discuss DRL applications in data preparation and then in data analytics.

\subsection{Data Preparation}

\subsubsection{Entity Matching}
Entity matching is a data cleaning task that aligns different mentions of the same entity in the context. Clark et al. \cite{clark-manning-2016-deep} identify the issue that the heuristic loss function cannot effectively optimize the evaluation metric $B^3$, and propose using reinforcement learning to directly optimize the metric. The problem is formulated as a sequential decision problem where each action is performed on one mention of a document. The action maps the mention to an entity in the database at each step by a mention ranking model. Then the reward is calculated using the evaluation metric $B^3$. This work originally proposes scaling each action's weight by measuring its impact on the final reward since each action is independent. However, this work does not consider the global relations between entities. Fang et al. \cite{DBLP.abs-1902-00330} propose a reinforcement learning framework based on the fact that an easier entity will create a better context for the subsequent entity matching. Specifically, both local and global representations of entity mentions are modeled and a learned policy network is devised to choose from the next action (i.e., which entity to recognize). However, the selection of the easier entity to learn the context could be less powerful than context modeling with more recent techniques in NLP such as the transformer.

\subsubsection{Database Interaction With Natural Language}
To facilitate query formulation for relational databases, there have been efforts in generating SQL queries from various other means that do not require knowledge of SQL and schema. Zhong et al. \cite{zhong2017seq2sql} propose to generate SQL from a natural language using Reinforcement Learning. For queries formed by a natural language, the model Seq2SQL will learn a policy transforming the queries into SQL queries. The transformed queries will then be executed in the database system to get results. The results will be compared with the ground truth to generate RL rewards. Earlier work~\cite{dong-lapata-2016-language} using generic autoencoder model for semantic parsing with Softmax as the final layer may generate unnecessarily large output spaces for SQL query generation tasks. Thus the structure of SQL is used to prune the output space of query generating and policy-based reinforcement learning to optimize the part which cannot be optimized by cross-entropy. However, RL is observed to have limited performance enhancement by \cite{xu2017sqlnet} due to unnecessary modeling of query serialization.

Efficiently querying a database of documents is a promising data processing application. Karthik et al. \cite{narasimhan-etal-2016-improving} propose collecting evidence from external sources of documents to boost extraction accuracy to original sources where data might be scarce. The problem is formulated as an MDP problem, where each step the agent needs to decide if current extracted articles are accepted and stop querying, or these articles are rejected and more relevant articles are queried. Both data reconciliation (from original sources) and data retrieval (from external sources) are represented as states. Extraction accuracy and penalties for extra retrieval actions are reflected in the reward function.

\subsubsection{Feature Engineering}
Feature engineering can be formulated as a single-agent reinforcement learning problem to search for an optimal subset of features in a large space: the agent selects one feature at each action step. The state is the current feature subspace. A reward is assigned to the agent based on the predictive performance of the current features subset. Liu et al. \cite{10.1145/3292500.3330868} propose a method to reformulate feature engineering as a multi-agent reinforcement learning problem. The multi-agent RL formulation reduces the large action space of a single agent since now each of the agents has a smaller action space for one feature selection. However, this formulation also brings challenges: interactions between agents, representation of the environment, and selection of samples. Three technical methods in \cite{10.1145/3292500.3330868} have been proposed to tackle them respectively: adding inter-feature information to reward formulation, using meta statistics, and deep learning methods to learn the representation of the environment, and Gaussian mixture to independently determine samples. However, although this formulation reduces the action space, the trade-off is using more computing resources to support more agents' learning. Also, the method is difficult to scale to a large feature space.

\subsubsection{Exploratory Data Analysis}
Exploratory data analysis (EDA) is useful for users to understand the characteristics of a new dataset. In~\cite{10.1145/3318464.3389779}, the problem is formulated as a MDP. The action space is the combination of a finite set of operators and their corresponding parameters to query a dataset. The result of a query shows the characteristics of the dataset. The characteristics are modeled as the state, which is represented by descriptive statistics and recent operators. The reward signal measures the interestingness, diversity, and coherency of the characteristics by an episode of EDA operations. DRL is applied to the non-differential signals and discrete states in MDP. However, challenges arise when applying deep reinforcement learning given a large number of possible actions as parameterized operations (i.e., for each type of operation, the corresponding possible action is the Cartesian product of all parameters' possible values). In~\cite{10.1145/3318464.3389779}, a two-fold layer architecture is proposed to replace a global softmax layer into two local layers, which effectively reduces the intractable large numbers of actions. However, the global interactions of operations and attributes are not considered.

\subsubsection{Abnormal Detection}
Abnormal detection is important for high-stake applications such as healthcare (e.g., predicting patients' status) and fintech (e.g., financial crime). Based on the assumptions, there are two approaches to this problem. One approach models the dynamics in the unlabeled datasets as a sequential decision process where the agent performs an action on each observation. Oh et al. \cite{10.1145/3292500.3330932} propose to use IRL to learn a reward function and a Bayesian network to estimate a confidence score for a potential abnormal observation. To achieve this, the prior distribution of the reward function is assumed. Then a reward function is sampled from the distribution to determine the sample generating policy, which generates sample background trajectories. As explained by the reward part of Section \ref{Model Efficiency}, experts' trajectories are observed. With these experts' trajectories and sample background trajectories, the parameters of the reward function are updated and thus the policy is improved. The sequence of actions is the input into the neural network. This network is trained to learn the normal pattern of a targeted agent and to predict if the next observation is abnormal or not. However, this approach relies too much on mining unlabeled datasets and ignores the labeled dataset. To address this issue, another approach also uses DRL but focus on the Exploit-Explore trade-off on both unlabeled and labeled dataset. Pang et al. \cite{Pang2020TowardDS} propose a DRL model with a sampling function to select data instances from both the unlabeled and labeled dataset. This sampling function helps the DRL model to exploit the scarce but useful labeled anomaly data instances and to explore the large unlabeled dataset for novel anomaly data instances. Thus, more anomaly data instances are selected to train the DRL model with better model capacity.

\subsubsection{AutoML Pipeline Generation}
Pipeline generation includes generating all data processing and analytics steps or modules to perform ML tasks. Heffetz et al. \cite{10.1145/3394486.3403261} propose a grid-world to represent all possible families of each step of a data pipeline as cells and connect all possible cells as a graph. Subsequently, a hierarchical method is used to reduce the space of all actions and represent all actions by layers of clusters. Finally, the state representations are inputs to the value sub-network in a DQN network, and action representations are inputs to evaluate the advantage-to-average sub-network.

\subsection{Healthcare}
Healthcare analytics has gained increasing attention in tandem with the advancement of healthcare treatment and availability of medical data and computational capacity \cite{Lee2017BigHD}. Naturally, a great amount of effort has been spent on applying DRL to healthcare. As before, implementing DRL-based models in healthcare requires the understanding of the application context and defining the key elements of MDP. However, differences occur in the approaches to learning better decisions: learning the motivation of expert decisions by IRL, learning better decisions without an expert by interacting with an environment or interacting with an environment with expert decisions as supervising signals.

\subsubsection{Treatment Recommendation}
Treatment recommendation systems are designed to assist doctors to make better decisions based on electronic health records. However, the doctors' prescriptions are not ground truth but valuable suggestions for high stake medical cases. The ground truth is the delayed condition of the patients. Thus model predictions must not deviate from the doctors' judgments too much, and not use those judgments as true labels. To tackle this challenge, Wang et al. \cite{10.1145/3219819.3219961} propose an architecture to combine supervised learning and reinforcement learning. This model reduces the inconsistency between indicator signals learned from doctor's prescriptions via supervised learning and evaluation signals learned from the long-term outcome of patients via reinforcement learning. In the formulated MDP, the domain expert makes a decision based on an unknown policy. The goal is to learn a policy that simultaneously reduces the difference between the chosen action of the agent and the expert's decision and to maximize the weighted sum of discounted rewards.

\subsubsection{Diagnostic Inference}
Using DRL to perform diagnosis can provide a second opinion in high-intensity diagnosis from historical medical records to reduce diagnostic errors. Ling et al. \cite{Ling2017DiagnosticIV} propose modeling the integration of external evidence to capture diagnostic concept as a MDP. The objective is to find the optimal policy function. The inputs are case narratives and the outputs are improved concepts and inferred diagnoses. The states are a set of measures over the similarity of current concepts and externally extracted concepts. The actions are whether to accept (part of) the extracted concepts from external evidence. The environments are the top extracted case narratives from Wikipedia as the document pool for concepts extraction and a knowledge base for evaluating the intermediate results for current best concepts. The rewards are evaluated based on an external knowledge base mapping from the concepts to the diagnoses. The whole process is modeled by DQN. At each step, narrative cases and evidence are extracted, which provide the initial concepts and external concepts. The state representing the agent's confidence in the learned concept is duly calculated. Then the state is sent to the DQN agent to estimate the reward to model the long-run accuracy of the learned concept by the agent. Iteratively, the model converges with better concepts and diagnoses.

\subsubsection{Hospital Resource Allocation}
Allocating limited hospital resources is the key to providing timely treatment for patients. In \cite{pmlr-v119-el-bouri20a}, the problem is formulated as a classification problem where the patients' features are given and the target is to predict the location of admissions. The RL framework uses a student network to solve the classification problem. The weights of the student network are used as states, which are fed into a teacher network to generate actions to select which batch of data to train the student network. The accuracy of the classification is used as the reward. This method provides a view on the resource allocation problem from a curriculum learning perspective. However, the temporal information of the data samples is not considered but it could affect resource allocation since some hours during a day could have fewer patients than the others.

\subsection{Fintech}

\label{fintech}
\begin{figure}

\centering

\includegraphics[width=0.8\textwidth]{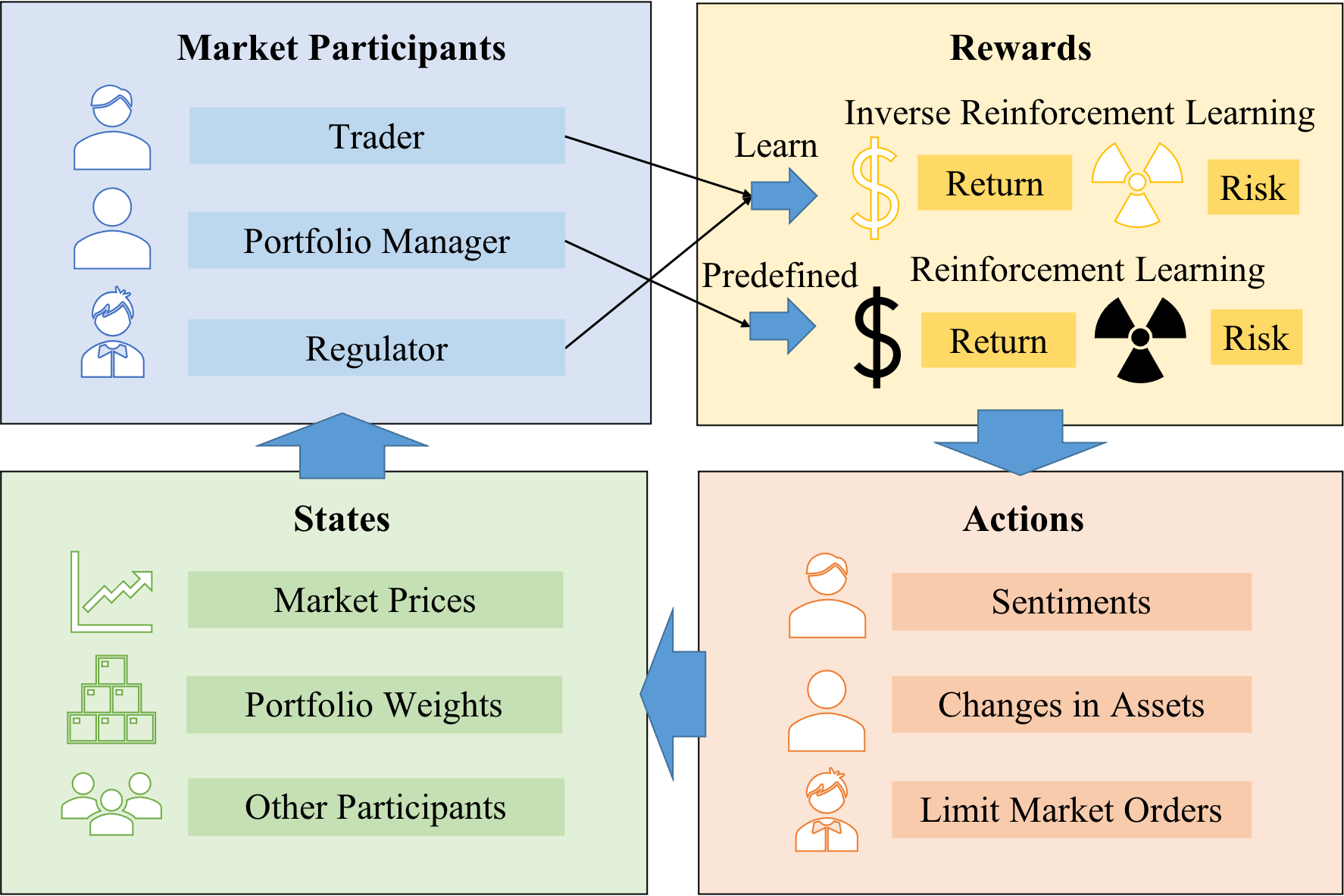}

\centering

\caption{DRL in fintech applications.}

\label{fig_fintech}

\end{figure}

Reinforcement learning has wide applications in the finance domain. Firstly, reinforcement learning has brought new perspectives to let the finance research community revisit many classic financial research topics. For example, traditional financial research topics such as option pricing that are typically solved by the classic Black–Scholes model can be steered through with a data-driven insight by reinforcement learning \cite{doi:10.1080/14697688.2019.1622302}. Secondly, portfolio optimization, typically formulated as a stochastic optimal control problem, can be addressed by reinforcement learning. Finally, the agents are financial market participants with different intentions. Reward functions can be learned to model these intentions, and hence, make better decisions as illustrated in Figure \ref{fig_fintech}. We refer readers with further interest in finance to \cite{coqueret2021machine}.

\subsubsection{Dynamic Portfolio Optimization}
The portfolio optimization problem is challenging because of the high scale of the dimensionality and the high noise-to-signal ratio nature of stock price data. The latter problem of noisy observation can cause uncertainty in a learned policy. Therefore, \cite{dixon2020glearner} proposes a novel model structure based on the Q-learning to handle noisy data and to scale to high dimensionality. The quadratic form of reward function is shown to have a semi-analytic solution that is computationally efficient. In the problem formulation, the agent's actions are represented as the changes in the assets at each time step. The states are the concatenation of market signals and the agent's holding assets. This method enhances Q-learning by introducing an entropy term measuring the noise in the data. This term acts as a regularization term forcing the learned policy to be close to a reference policy that is modeled by a Gaussian distribution.
\subsubsection{Algorithm Trading Strategy Identification}
Identification of algorithm trading strategies from historical trades is important in fraud detection and maintaining a healthy financial environment. \cite{articleYang} proposes using IRL to learn the reward function behind the trading behaviors. The problem is formulated as an Inverse Markov Decision Process (IMDP). The states are the differences between the volumes of bid orders and ask orders, which are discretized into three intervals based on the values of the volumes. The actions are the limit and market order discretized into 10 intervals each by their values. The prior distribution of the reward function is a Gaussian Process parameterized by $\theta$. Given $\theta$, the approximation of the posterior distribution of reward is performed by maximum a posteriori (MAP). This step would give a MAP estimated value of the reward. $\theta$ is optimized by a log-likelihood function on the posterior of observations. The optimization process can be proved to be convex which guarantees the global minimum. The learned features are then used to identify and classify trading strategies in the financial markets.

\subsubsection{Sentiment-based Trading}
One of the main predictors in stock trading is sentiment, which drives the demand of bid orders and asks orders. Sentiment scores are often represented by unstructured text data such as news or twitters.
\cite{YANG2018388} proposes treating the sentiment as the aggregated action of all the market participants, which has the advantage of simplifying the modeling of the numerous market participants. Specifically, the sentiment scores are categorized into three intervals: high, medium, and low as the action spaces. Compared to previous works, the proposed method can model the dependency between the sentiment and the market state by the policy function.
This method is based on Gaussian Inverse Reinforcement Learning \cite{NIPS2011_c51ce410} similar to \cite{articleYang} as discussed at the beginning of Section \ref{fintech}, which is effective at dealing with uncertainty in the stock environment. This method provides a method for modeling market sentiments. However, as IRL faces the challenge of non-uniqueness of reward \cite{dong2020deep} of one agent's actions, the method does not address how aggregated actions of multiple market participants can infer a unique reward function.

\subsection{E-Commerce}

\subsubsection{Online Advertising}
With the increasing digitalization of businesses, sales and competition for market shares have moved online in tandem. As a result, online advertising has been increasing in its presence and importance and exploiting RL in various aspects. One of the topics in online advertising, bidding optimization, can be formulated as a sequential decision problem: the advertiser is required to have strategic proposals with bidding keywords sequentially to maximize the overall profit. In~\cite{10.1145/3219819.3219918}, the issue of using static transitional probability to model dynamic environments is identified and a new DRL model is proposed to exploit the pattern discovered from dynamic environments.

Including but not limited to advertising, Feng et al. \cite{10.1145/3178876.3186165} propose to consider the whole picture of multiple ranking tasks that occurred in the sequence of user's queries. A new multi-agent reinforcement learning model is proposed to enable multiple agents to partially observe inputs and choose actions through their own actor networks. The agents communicate through a centralized critic model to optimize a shared objective. This allows different ranking algorithms to reconcile with each other when taking their own actions and consider the contextual information.

\subsubsection{Online Recommendation}
The problem of an unstabilized reward function arises because of the dynamic environment in the online recommendation. For example, user preference is modeled as the reward in DRL and it changes unexpectedly when a special discount happens for some products. In \cite{10.1145/3219819.3220122}, a random stratified sampling method is proposed to calculate the optimal way of stratifying by allocating more samples to the strata with more weighted variance. Then the replay sampling is improved to consider key attributes of customers (e.g., gender, age, etc.), which are less volatile in the dynamic environment. This allows the modeling of reward function based on sampling from a pool with a longer horizon, thus reducing the bias in the estimation of the reward function. Lastly, the dynamic environment poses a challenge in setting an optimal policy used in regretting. A new method in \cite{10.1145/3219819.3220122} is proposed to train an offline model to calculate a real-time reward for a subset of customers to approximate a reference policy, that is used as an offset in the reward recalibration to stabilize the performance of the DRL algorithm.

\subsubsection{Search Results Aggregation}
Aggregating useful search results in online shopping search is important to improve the shopping experience. However, the challenge of aggregating heterogeneous data sources is often encountered. The heterogeneous data sources in online shopping are different product categories such as a shoe brand group or a particular topic group, each of which is a ranking system. A new model in \cite{10.1145/3308558.3313455} is proposed to decompose the task into two sub-tasks. The first one is to select a data source for the current page of search results based on historical users' clicks on previous pages. Learning to select the correct data source for each page is a sequential decision-making problem. The second sub-task is to fill the sequence of a page by selecting the best source from the candidate sources. However, the items from different sources cannot be directly compared because of their heterogeneous nature. The problem is solved by formulating the sub-task as an RL task to let an agent fill up the sequence. However, one limitation of this method is that lacking full annotations of item relevance scores may constrain the model's performance on various scenarios \cite{10.1145/3308558.3313455}.

\subsection{Other Applications}
DRL has been applied to various other applications. These DRL methods are often used with a knowledge graph, confounders, or game theory to model application-specific dynamics. These methods are not only well motivated from their respective applications but also general enough to be applied in other applications. However, these methods often fail to be evaluated by experiments in other applications.

The problem of mobile user profiling aims to identify user profiles to provide personalized services. In \cite{10.1145/3394486.3403128}, the action is the selection of a place of visit. The environment is comprised of all users and a knowledge graph learning the semantic connections between the spatial entities. The knowledge graph is updated once a user's new activity is performed and then affects the agent's prediction. The state is the embedding of a user and the knowledge graph for the current time step. The reward is determined by several metrics measuring the similarity between the predicted spatial entities and the ground truth. This method considers the spatial semantics of entities but does not consider how the change of a user's key attributes (e.g., career) will affect activity prediction and policy learning, which could cause instability in policy updating.

In the transportation system, drivers often get recommendations and provide feedback in return to improve the service. However, the recommendation often fails when drivers make decisions in a complex environment. To address this issue, in \cite{10.1145/3292500.3330933} a new method is proposed to model hidden causal factors, called confounders, in a complex environment. Specifically, the framework in \cite{ho2016generative} is extended to include the confounders. First, all three elements (i.e., policy agent, environment, confounder) are treated as agents. The effect of a confounder is modeled as the policy of the hidden agent, which takes the observation and action of the policy agent as inputs and performs an action. The environment in turn takes the action based on inputs of the hidden agent's action and the policy agent's action and observation.

The problem of spammer detection aims to detect spam generating strategies. The challenge is that the detectors only detect easier spams while missing spams with strategies. In \cite{10.1145/3394486.3403135}, the problem is formulated as two agents counteracting each other. One agent is the spammer, whose policy is to maintain a distribution of spam strategies and the action is to sample from the distribution. Another agent is the detector, whose state is the detection results after a spam attack and the action is to identify the spam. The rewards of two agents are measured by winning or losing revenue manipulation, respectively. The limitation of this method is that there is no guarantee for equilibrium.

\section{Open Challenges and Future Directions}
RL approaches provide strong alternatives to traditional heuristics or supervised learning-based algorithms. However, many challenges remain to be addressed to make RL a practical solution in the context of data processing and analytics. We also foresee many important future research directions to be developed.

\subsection{Open Challenges For System Optimization}

\subsubsection{MDP Formulation and Lack of Justification}
The design of MDP impacts the performance and efficiency of the RL algorithm greatly. The state should satisfy Markov property that its representation contains enough relevant information for the RL agent to make the optimal decision. It should summarize the environment compactly because a complicated state design will cause more training and inference costs. The action space should be designed carefully to balance learning performance and computational complexity. The reward definition directly affects the optimization direction and the system performance. Additionally, the process of reward calculation can involve costly data collection and computation in the data systems optimization. Currently, many works rely on experimental exploration and experience to formulate MDP while some works exploit domain knowledge to improve the MDP formulation by injecting task-specific knowledge into action space\cite{welborn2019learning}. Generally, MDP can influence computational complexity, data required, and algorithm performance. Unfortunately, many works lack ablation studies of their MDP formulations and do not justify the design in a convincing manner. Therefore, automation of MDP formulation remains an open problem.

\subsubsection{RL Algorithm and Technique Selection}
RL algorithms and techniques have different tradeoffs and assumptions. Value-based DRL algorithms like DQN are not stable and guaranteed convergence. Policy-based DRL algorithms like TRPO and PPO are often not efficient. Model-based DRL algorithms do not guarantee that a better model can result in a better policy. Value-based methods assume full observability while policy-based ones assume episodic learning. Off-policy algorithms are usually more efficient than on-policy algorithms in terms of sample efficiency. One example is that DQ\cite{krishnan2018learning} uses off-policy deep Q-learning to increase data efficiency and reduce the number of training queries needed. Training efficiency can be a big concern for DRL-based system optimization, especially when the workload of the system could change dramatically and the model needs to be retrained frequently. Generally, RL algorithms and techniques selection affect the training efficiency and effectiveness greatly.

\subsubsection{Integration with Existing Systems}
Integrating RL-based methods into the real system more naturally and seamlessly faces many challenges. The RL agent has to be evolved when the system environment changes (e.g., workload) and the performance is degraded. We need to design new model management mechanisms to monitor, maintain, and upgrade the models. Furthermore, we find that the RL-based solutions can be lightweight or intrusive. The lightweight approach in which the RL agent is not designed as a component of the system, e.g. using RL to generate the qd-tree\cite{yang2020qd}, is easier to integrate into the system because it does not change the architecture of the system dramatically. In contrast, the intrusive approach such as using RL models for join order optimization\cite{marcus2018deep} is deeply embedded in the system and hence may need a redesign and optimization of the original system architecture to support model inference efficiently. SageDB\cite{kraska2019sagedb} proposes to learn various database system components by integrating RL and other ML techniques. Nevertheless, the proposed model-driven database system is yet to be fully implemented and benchmarked. It is likely that the data system architecture needs to be overhauled or significantly amended in order to graft data-driven RL solutions into the data system seamlessly to yield an overall performance gain.

\subsubsection{Reproducibility and Benchmark}
In the data system optimization problem, RL algorithms are not easy to be reproduced due to many factors such as lacking open source codes, workload, historic statistics used, and the unstable performance of RL algorithms. The landscape of problems in system optimization is vast and diverse. It could prevent fair comparison and optimization for future research works and deployments in practice. Lacking benchmarks is another challenge to evaluate these RL approaches. The benchmarks are therefore to provide standardized environments and evaluation metrics to conduct experiments with different RL approaches. There are some efforts to mitigate the issue. For example, Park\cite{mao2019park} is an open platform for researchers to conduct experiments with RL. However, it only provides a basic interface and lacks system specifications. There is much room to improve with regards to the reproducibility and benchmark in order to promote the development and adoption of RL-based methods\cite{henderson2018deep}.

\subsection{Open Challenges For Applications}

\subsubsection{Lack of Adaptability}
There is a lack of adaptability for methods on a single component of a data pipeline to the whole. For example, many works focus on data cleaning tasks such as entity matching. However, little works have shown their efficiency in deploying their model in an end-to-end data pipeline. These works treat the tasks isolatedly from other tasks in the pipeline, thereby limiting the pipeline's performance. In healthcare, each method is applied in different steps of the whole treatment process, without being integrated and evaluated as one pipeline. One possible direction could be considering DRL as a module in the data pipeline optimization. However, data pipeline optimization has been focusing on models simpler than DRL to enable fast pipeline evaluation \cite{9458924}. How to efficiently incorporate DRL into the data pipeline optimization remains a challenge.

\subsubsection{Difficulty in Comparison with Different Applications}
To date, most works with generalized contributions are only evaluated domain-specifically. Research questions are often formulated in their own platform as in E-Commerce. This presents difficulty in evaluating the methods for different environments. For example, the confounders modeling hidden causal factors in \cite{10.1145/3292500.3330933} can also contribute to DRL modeling in E-commerce. This is because modeling customers' interests are always subject to changing environments and a new environment may contain hidden causal factors. For example, consumers are more willing to buy relevant products for certain situations such as Covid-19. Thus a general DRL method is yet to show the robustness and effectiveness under the environment of different applications.

\subsubsection{Lack of Prediction in Multi-modality}
In healthcare and finance, multiple sources of data bring different perspectives. For example in healthcare, electronic health records, image scans, and medical tests can provide different features for accurate prediction. In addition, these sources of data with different sample frequencies provide contextual information for modeling a patient's visits to the hospital or symptom development. However, most innovations in healthcare focus on one particular source of data. How to integrate the contextual information with multi-modality effectively remains an unsolved difficult problem.

\subsubsection{Injecting Domain Knowledge in Experience Replay}
In high-stake applications such as healthcare and finance, injecting domain knowledge can make decision making in RL more robust and explainable. One possible way is to inject the knowledge of human beings' experience into an agent's experience pool as a prior distribution for the policy. For example, in dynamic portfolio optimization, a portfolio manager could have a large source of experience for risk management and profit optimization. Such experience could be useful for warming up the agent's exploration in the search space. Some works have shown positive effects of domain knowledge injection on selecting important experiences (i.e., transition samples) \cite{schaul2016prioritized}. Notwithstanding, it remains a big challenge to inject useful and relevant knowledge from the experience into the agent's experience pool.

\subsection{Future Research Directions}

\subsubsection{Data Structure Design}
DRL provides an alternative way to find good data structures through feedback instead of designing them based on human knowledge and experience, e.g., decision tree\cite{liang2019neural} and the qd-tree\cite{yang2020qd}. These trees are optimized better because they are learned by interacting with the environment. DRL has also been effective in graph designs (e.g., molecular graph\cite{you2018graph}). However, large-scale graph generation using DRL is difficult and daunting because it involves a huge search space. Generating other important structures using DRL remains to be explored. Idreos et al.\cite{idreos2019learning} propose a \textit{Data Alchemist} that learns to synthesize data structures by DRL and other techniques including Genetic Algorithms and Bayesian Optimization. In summary, DRL has a role in the design of more efficient data structures by interacting and learning from the environment. These indexes have to be adaptive to different data distributions and workloads.

\subsubsection{Interpretability}
The underlying logic behind the DRL agent is still unknown. In high-risk application areas such as healthcare, the adoption of DRL will be a big issue in the case that these approaches make wrong decisions and people do not know why it happens due to lack of interpretability. Many techniques have been proposed to mitigate the issue and provide interpretability\cite{puiutta2020explainable}. However, they neglect domain knowledge from related fields and applications and the explanations are not effective to human users. To instill confidence in the deployment of DRL-based systems in practice, interpretability is an important component and we should avoid treating DRL solutions as black boxes especially in critical applications.

\subsubsection{Robustness by Causal Reasoning}
Modeling real-world applications by DRL inevitably suffers from the problem of distribution changes. The real world has independent physical mechanisms that can be seen as different modules. For example, an image is subjected to the light of the environment. Given the modular property, a structural type of modeling focusing on factorizing the causal mechanisms can extract the invariant causal mechanisms and show robustness cross distribution changes \cite{9363924}. One research direction towards DRL robust decision making is to perform sampling from past actions from a causal perspective. Given the invariance property of causal mechanisms, past actions can be reused by capturing the invariant mechanisms in a changing environment.

\subsubsection{Extension to Other Domains}
Beyond existing works, many classic problems in the data system and analytics could potentially be solved by DRL. For example, Polyjuice\cite{273747} learns the concurrency control algorithm for a given workload by defining fine-grained actions and states in the context of concurrency control. Though they use an evolutionary algorithm to learn and outperform a simple DRL baseline, we believe that there are huge potentials to further improve DRL for niche applications. Hence, we expect that more problems will be explored and solved with DRL in various domains in the near future.

\subsubsection{Towards Intelligent and Autonomous Databases}
Although DRL algorithms could provide breakthrough performance on many tasks than traditional methods, many issues need to be addressed towards intelligent and autonomous databases. First, database schema could be updated and DRL models trained on the previous snapshots may not work. DRL algorithms need to tackle generalization\cite{packer2018assessing}. Second, it would be so costly and infeasible to train models from scratch for each scenario and setting. Transfer learning from existing models could be a potential way to ease the workload greatly. Third, we have to choose appropriate DRL algorithms automatically, in the same spirit as AutoML. Fourth, current DBMS systems were designed without considering much about the learning mechanism. A radically new DBMS design may be proposed based on the learning-centric architecture. To support intelligent and autonomous database systems, DRL models intelligent behaviors and may provide a solid basis for achieving artificial general intelligence based on reward maximization and trial-and-error experience\cite{silver2021reward}.

\section{Conclusions}
In this survey, we present a comprehensive review of recent advances in utilizing DRL in data processing and analytics. The DRL agent could learn to understand and solve various tasks with the right incentives. First, we introduce basic foundations and practical techniques in DRL. Next, we survey and review DRL for data processing and analytics from two perspectives, systems and applications. We cover a large number of topics ranging from fundamental problems in system areas such as tuning and scheduling to important applications such as healthcare and fintech. Finally, we discuss key challenges and future directions for applying DRL in data processing and analytics. We hope the survey would serve as a basis for research and development in this emerging area, and better integration of DRL techniques into data processing pipelines and stacks.


\begin{spacing}{0.99}
\bibliographystyle{abbrv}
\bibliography{ref}
\end{spacing}

\end{document}